\documentclass{article}
\usepackage{makecell}
\usepackage{microtype}
\usepackage{graphicx}
\usepackage{subfigure}
\usepackage{booktabs} % for professional tables
\usepackage{multirow}

\usepackage{hyperref}
\usepackage[dvipsnames,svgnames, table,xcdraw]{xcolor}
\definecolor{mediumpurple}{rgb}{0.58, 0.44, 0.86}

\usepackage{amsmath}
\usepackage{amssymb}
\usepackage{mathtools}
\usepackage{amsthm}
\usepackage{enumerate}

\usepackage[normalem]{ulem}
\useunder{\uline}{\ul}{}

\usepackage{blindtext}

\usepackage[T1]{fontenc}
\usepackage{subfiles} 
\usepackage{amsmath} 
\usepackage{amsthm}
\usepackage{enumerate}
\usepackage{graphicx}
\usepackage{subcaption}
\usepackage{subfigure}

\usepackage{pdfpages}
\usepackage{amsfonts}
\usepackage{paralist}
\usepackage{booktabs} % To thicken table lines

\usepackage{amsmath,amsfonts,bm}

\def\eqref#1{equation~\ref{#1}}
\def\1{\bm{1}}

\DeclareMathAlphabet{\mathsfit}{\encodingdefault}{\sfdefault}{m}{sl}
\SetMathAlphabet{\mathsfit}{bold}{\encodingdefault}{\sfdefault}{bx}{n}

\usepackage[accepted]{icml2025}

\usepackage{amsmath}
\usepackage{amssymb}
\usepackage{listings}
\usepackage{tcolorbox}
\usepackage{mathtools}
\tcbuselibrary{listingsutf8} 
\usepackage{xcolor}
\usepackage{amsthm}
\definecolor{codeborder}{RGB}{50, 100, 200}  
\definecolor{codebackground}{RGB}{245, 245, 245}  
\definecolor{titlebg}{RGB}{220, 230, 250} 
\definecolor{titleborder}{RGB}{50, 100, 200}

\newtcolorbox[auto counter, number within=section]{codebox}[2][]{%
    colback=codebackground, 
    colframe=codeborder,  
    coltitle=black, 
    colbacktitle=titlebg,  
    coltitle=Black,  
    fonttitle=\bfseries,  
    boxrule=1.2pt, 
    arc=5pt,  
    listing only, 
    listing options={
        language=Python,
        basicstyle=\ttfamily\footnotesize,
        keywordstyle=\color{blue},
        stringstyle=\color{red},
        commentstyle=\color{gray},
        showstringspaces=false,
        columns=flexible,
        breaklines=true,
        inputencoding=utf8,  
        extendedchars=true
    },
    title=Python Code: #2,  
    #1
}

\usepackage{smile}
\usepackage{cases}

\newcommand{\model}{\textcolor{RoyalBlue}{\textsc{SiriuS}}}

\usepackage[capitalize,noabbrev]{cleveref}
\theoremstyle{plain}

\theoremstyle{definition}

\theoremstyle{remark}

\usepackage[textsize=tiny]{todonotes}

\icmltitlerunning{\textcolor{RoyalBlue}{SiriuS}: Self-improving Multi-agent Systems via Bootstrapped Reasoning}

\begin{document}

\twocolumn[
\icmltitle{\textcolor{RoyalBlue}{SiriuS}: \textcolor{RoyalBlue}{S}elf-\textcolor{RoyalBlue}{i}mp\textcolor{RoyalBlue}{r}ov\textcolor{RoyalBlue}{i}ng M\textcolor{RoyalBlue}{u}lti-agent \textcolor{RoyalBlue}{S}ystems via Bootstrapped Reasoning}

\vspace{-10pt}
\begin{icmlauthorlist}
\icmlauthor{Wanjia Zhao}{yyy}
\icmlauthor{Mert Yuksekgonul}{yyy}
\icmlauthor{Shirley Wu}{yyy}
\icmlauthor{James Zou}{yyy}
\end{icmlauthorlist}

\icmlaffiliation{yyy}{Stanford University}
\icmlcorrespondingauthor{Wanjia Zhao}{wanjiazh@cs.stanford.edu}
\icmlcorrespondingauthor{James Zou}{jamesz@cs.stanford.edu}

\vskip 0.3in
]

\printAffiliationsAndNotice{} % otherwise use the standard text.

\begin{abstract}
Multi-agent AI systems powered by large language models (LLMs) are increasingly applied to solve complex tasks. However, these systems often rely on fragile, manually designed prompts and heuristics, making optimization difficult.
A key challenge in optimizing multi-agent systems is acquiring suitable training data for specialized agents. 
We introduce \model{}, a self-improving, reasoning-driven optimization framework for multi-agent systems. Central to our approach is the construction of an experience library: a repository of high-quality reasoning trajectories. The library is built by retaining reasoning steps that lead to successful outcomes, providing a robust training set for optimizing multi-agent system. Additionally, we introduce a library augmentation procedure that refines unsuccessful trajectories, further enriching the library. 
\model{} boosts performance by 2.86\% to 21.88\% on reasoning and biomedical QA and enhances agent negotiation in competitive settings. Our results show that \model{} enhances multi-agent performance while generating reusable data for self-correction and self-play enhancement in the future. Code are available \href{https://github.com/zou-group/sirius}{here}.

\end{abstract}

\section{Introduction} 
\begin{figure*}[htbp]
    \centering
    \includegraphics[width=1\linewidth]{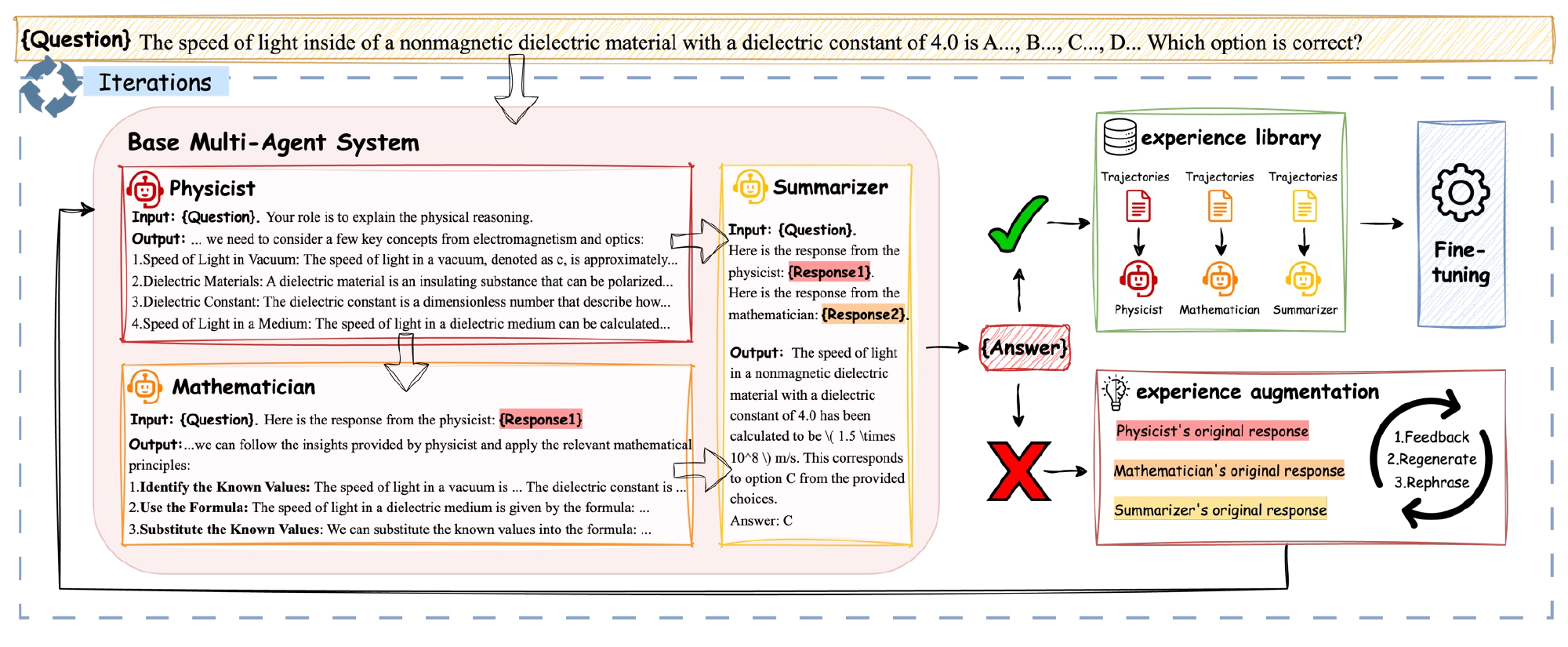}
    \caption{General training pipeline of \model.Agents solve problems sequentially, storing correct responses for fine-tuning and augmenting incorrect ones through feedback, regeneration, and rephrasing. This iterative process improves performance via reward-based evaluation and supervised fine-tuning. The module colors in the figure correspond to those in Algorithm \ref{alg:method}.
}
    \label{fig:pipeline}
\end{figure*}

Multi-agent AI systems powered by large language models~(LLMs), where specialized agents collaborate to solve complex tasks, are becoming increasingly successful in real-world applications. Recent work has demonstrated their effectiveness in complex reasoning~\citep{wang2024mixture, smitshould}, coding~\citep{wu2023autogen}, drug discovery~\citep{swanson2024virtual} and ensuring safety via debate~\citep{chern2024combating, irving2018ai}. 
These successes arise from specialized agents integrating their distinct capabilities through structured interactions, enabling more effective problem-solving than single agents. Moreover, multi-agent scrutiny acts as a built-in self-correction mechanism, where agents refine and verify each other’s outputs. This often outperforms single agent setting, particularly on tasks demanding rigorous reasoning or factual validation.

Despite these successes, optimizing multi-agent systems remains a fundamental challenge due to (1) the difficulty of acquiring appropriate training signals for each agent and (2) the sensitivity to multiple moving parts that influence overall performance~\citep{smitshould}. While task-level reward feedback is available, credit assignment across agents remains ambiguous—it is unclear how to attribute success or failure to specific intermediate decisions and reasoning steps made by each LLM agent. This challenge parallels the multi-agent credit assignment problem in reinforcement learning~\citep{foerster2018counterfactual}. However, in language-based systems, reasoning unfolds through complex and unstructured interactions, making attribution far more difficult than in traditional RL settings with well-defined action spaces.

We present \model{}, a framework for learning effective multi-agent behaviors from outcome rewards. Our key insight is that when multiple agents successfully solve a task together, their entire interaction trajectory likely contains useful patterns - even if we cannot pinpoint exactly which steps or decisions were crucial for success. Drawing inspiration from recent advances in bootstrapping reasoning capabilities~\citep{zelikman2022star}, we collect and learn from successful agent interactions across many tasks, allowing the system to iteratively discover effective collaboration strategies from self-generated data. This approach sidesteps the need for direct supervision of intermediate steps, instead letting agents learn which interaction patterns tend to lead to successful outcomes. For trajectories that result in failed attempts, we perform trajectory augmentation by resampling original attempts with feedback from an additional agent grounded in the ground truth.

Our experiments demonstrate that \model{} significantly enhances multi-agent performance across multiple domains. It improves reasoning and biomedical QA accuracy by 2.86\% to 21.88\%, while also strengthening agent negotiation in competitive scenarios. Beyond these gains, our approach offers a scalable mechanism for self-improvement, enabling agents to iteratively refine their reasoning and collaboration strategies. More broadly, \model{} provides a general framework for optimizing multi-agent systems via self-generated synthetic data, offering a principled way to enhance performance without requiring fine-grained human supervision.

\section{Method}

\begin{table*}[t]
\centering
\caption{
    Different settings and tasks.
    In the rows corresponding to Communication Structure, nodes denote agents ($\mathcal{V}$), arrows represent edges ($E$), and color indicates the role of agents.
    }
\resizebox{\textwidth}{!}{
\begin{tabular}{l||c|c||c||c}
\toprule
\textbf{Settings} & \multicolumn{2}{c||}{\textbf{Problem-Solving}} & {\textbf{Actor-Critic}} &{\textbf{Competitive}} \\ 
\midrule
\textbf{Structure $(\mathcal{V},E,\mathcal{P})$} & {\begin{minipage}[b]{0.45\columnwidth}
		\centering
		\raisebox{-.5\height}{\includegraphics[height=2.2cm]{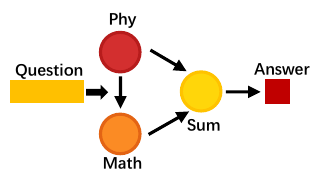}}
	  \end{minipage} }&\begin{minipage}[b]{0.4\columnwidth}
		\centering
		\raisebox{-.5\height}{\includegraphics[height=2.2cm]{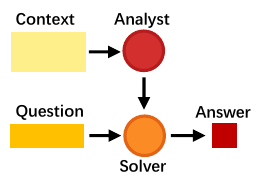}}
	  \end{minipage}& \begin{minipage}[b]{0.48\columnwidth}
		\centering
		\raisebox{-.5\height}{\includegraphics[height=2.2cm]{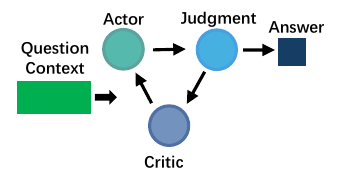}}
	  \end{minipage}&{\begin{minipage}[b]{0.4\columnwidth}
		\centering
		\raisebox{-.5\height}{\includegraphics[height=2.2cm]{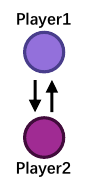}}
	  \end{minipage}} \\
\midrule
\textbf{Tasks} & {\makecell{College-Physics\\ College-Chemistry}}& PubMedQA & PubMedQA  &{\makecell{Resource Exchange\\ Seller-Buyer\\ Ultimatum}}  \\
\midrule
\textbf{Reward for each agent $R_i$} & \multicolumn{2}{c||}{Final Output Correctness} & \makecell{Final Output Correctness} &{Utility Function Value}\\ 
\bottomrule
\end{tabular}
}
    \label{tab:setting} 
\end{table*}

\subsection{Multi-agent systems with LLMs}  
%\shirwu{can we simplify this formulation a bit? For example, is $\gamma$ a necessary component?}
We define a multi-agent system by a tuple $\langle \mathcal{S}, \mathcal{A}, \mathcal{T}, \mathcal{R}, \mathcal{N}, \mathcal{G} \rangle$. Here, $\mathcal{N} \triangleq \{A^{(1)}, A^{(2)}, \ldots, A^{(N)}\}$ is the set of $N$ agents, each agent $A^{(i)}$ uses a policy $\pi_i$ parameterized by $\theta^{(i)}$. $s \in \mathcal{S}$ is the state of the environment, $\mathbf{a} \in \mathcal{A}$ is the joint actions, and $\mathcal{A}$ is the joint action space. $\mathcal{T}: \mathcal{S} \times \mathcal{A} \to \mathcal{S}$ is the transition function where $\mathcal{T}(s, \mathbf{a})$ yields the next state of the environment given the current state and joint actions $\mathbf{a}$. The environment feedback is modeled via a payoff function $\mathcal{R}_i: \mathcal{S} \times \mathcal{A} \to \mathbb{R}^N$, which provides rewards for each agent $k$ based on the state-action pairs. 

The communication structure between agents is modeled as a directed graph $\mathcal{G} = (\mathcal{V}, E, \mathcal{P})$, where $\mathcal{V}$ represents agents, and $E$ defines interaction order. 

For each edge $(i, j) \in E$, agent $A^{(j)}$ receives an input derived from the state-action pair $(s, \mathbf{a})$ and the output of agent $A^{(i)}$. This input determines agent $A^{(j)}$'s subsequent action. For each agent $A^{(i)}$ in a topological graph $\mathcal{G}$, its predecessors are the set of agents that  influence its output:
$\mathrm{Pre}(A^{(i)}) = \{A^{(j)} \mid (A^{(j)}, A^{(i)}) \in \mathcal{G}\}.$ Here, $(A^{(j)}, A^{(i)})$ denotes a directed edge in the graph, indicating that the output of agent $A^{(j)}$ directly influences the input of agent $A^{(i)}$.

Throughout this paper, the collection of our agents will be based on language models and the primary environment that we use will be natural language. In particular:

\begin{equation}
   \begin{aligned}
    & a_i \sim \pi_i(\cdot | s_{t}, \{a_j\}_{A^{(j)} \in \mathrm{Pre}(A^{(i)})}) \quad \forall A^{(i)}\in \mathcal{N} \\
    & \mathbf{a}_t = (a_1, ..., a_N) \\
    & s_{t+1} = \mathcal{T}(s_t, \mathbf{a}_t) = \text{Concat}(s_t, \mathbf{a}_t)
   \end{aligned}
\end{equation}

where $\pi_i$ denotes the probability distribution of the $i$-th language model, $\text{Concat}$ is the concatenation of the previous state and the responses, and we will use $\mathbf{\pi} = \{\pi_1, \ldots, \pi_N\}$ to denote the joint policy. Generally, each agent aims to maximize its own reward:
\begin{equation}
    \max_{\pi_i} \mathbb{E}_{\mathbf{\pi}}\left[\sum_{t=0}^{\infty}  R_i(s_t, \mathbf{a}_t)\right],
\end{equation}
where $R_i$ denotes the $i$-th component of the reward vector $\mathcal{R}$ and the expectation is taken under the joint policy $\mathbf{\pi}$. 

\subsection{\model}
The training pipeline of the proposed framework, denoted as \model, is illustrated in Figure~\ref{fig:pipeline}. \model{} adopts a fine-tuning strategy to iteratively improve the policy parameters $\theta^{(n)}$ of each agent $A^{(n)}$ over $T$ iterations. The process is initialized with a dataset $\mathcal{D} = \{(x_i, y_i)\}_{i=1}^D$, where each pair $(x_i, y_i)$ represents a problem and its solution.
The core training procedure is outlined in Algorithm~\ref{alg:method}. 
\begin{algorithm}[htbp]
\caption{\model}
\label{alg:method}
\begin{algorithmic}[1]
\STATE \textbf{Input:} A group of agents $A^{(1)},\cdots,A^{(N)}$.

An initial dataset of problems $x$ with answer $y:\mathcal{D} = \{(x_i, y_i)\}_{i=1}^D$, total number of fine-tuning Iterations $T$.
\STATE \textbf{Initialize:} Initialize policy parameters $\theta^{(n)}$ for each agent $A^{(n)}$, $k = 1, 2, \dots, N$.

\FOR{Fine-tuning Iteration $\text{t} =1,\cdots,T$}    
   
     \STATE \fcolorbox[HTML]{F8CECC}{F8CECC} {$a_i^{(n)}=\mathcal{P}_{\theta^{(n)}_{\text{t}}}(\cdot|x_i,\textbf{a}_i^{\mathrm{Pre}(A^{(n)})})$, $k = 1, 2, \dots, K$.}
        
        \FOR{each agent $k = 1, 2, \dots, K$}

            \STATE  \setlength{\fboxrule}{1.2pt}\fcolorbox[HTML]{82B366}{FFFFFF} {$\mathcal{C}_{\text{t}}^{(n)}\leftarrow \{(x_i, a_i^{(n)}| i \in [1,D] \land R_i(s,a)>\epsilon)\}$} Good Trajectory Set of Each Agent.
            \STATE \fcolorbox[HTML]{B85450}{FFFFFF}{ \textbf{Augmentation$(\{(x_i, a_i^{(n)} \land R_i(s,a)<\epsilon)\})$}} 
        \ENDFOR

    \STATE  \fcolorbox[HTML]{DAE8FC}{DAE8FC} {$\theta^{(n)}_{\text{t}} \leftarrow \textbf{Standard SFT on }\mathcal{C}_{\text{t}}^{(n)}$, $n=1,\cdots,N$}
\ENDFOR
\end{algorithmic}
\end{algorithm}

At each fine-tuning iteration $t$:
\begin{itemize}
    \item \textbf{Action Sampling:} For each agent \( A^{(n)} \), an action \( a_i^{(n)} \) is sampled from its policy, 
    \[
    a_i^{(n)} = \mathcal{P}_{\theta^{(n)}}(\cdot | x_i, \mathbf{a}_i^{\mathrm{Pre}(A^{(n)})}),
    \]
    conditioned on the input problem $x_i$ and the action set \( \mathbf{a}_i^{\mathrm{Pre}(A^{(n)})} \) generated by previous agents. In scenarios involving multiple interaction rounds, such as the Competitive Setting, \( \mathbf{a}_i^{\mathrm{Pre}(A^{(n)})} \) includes outputs from all agents in all preceding rounds.
    \item \textbf{Trajectory Evaluation and Augmentation:} The trajectories generated by each agent are evaluated using the payoff function $R(s, \mathbf{a})$. Based on a reward threshold $\epsilon$,  
    high-reward trajectories ($R(s, \mathbf{a}) > \epsilon$) are added to the good trajectory set $\mathcal{C}_{t}^{(n)}$. Since the tasks are challenging, the good trajectory set tends to be small. To leverage more data for fine-tuning, we propose trajectory augmentation pipeline for each task, detailed in the Appendix \ref{ap:pipeline}. Specifically, we first generate feedback to refine the agent's original response.
    The feedback and original response are then combined to prompt the agent to regenerate a new solution, which is then rephrased into a direct problem-solving step. Afterward, we return to the action sampling process to produce the final answer and evaluate it. 

    \item \textbf{Fine-Tuning:} The policy parameters $\theta^{(n)}$ are updated via supervised fine-tuning (SFT) on $\mathcal{C}_{t}^{(n)}$.
\end{itemize}

This iterative process ensures that each agent's policy is progressively refined to maximize performance based on the joint system dynamics and reward.

\section{Multi-agent Settings}
In this section, we explore several settings where agents with distinct expertise interact to solve challenging tasks. As shown in Table \ref{tab:setting}, we systematically analyze different agent configurations.

\begin{table*}[t]
\label{tab:competitive}
\caption{ Tasks and setups in the competitive setting. Each task involves two agents with distinct roles, initial resources, and objectives. \emph{Resource Exchange} focuses on maximizing total resources through trade. Ultimatum requires negotiating a split of $\$100$. \emph{Sell\&Buy} involves price negotiation for an item. Each task follows a turn-based structure with a fixed maximum number of rounds and ends when an agreement is reached.}
\centering
\small
\begin{tabular}{c|c|c|c|c|c|c}
\toprule
\textbf{Task} & \multicolumn{2}{c|}{\textbf{Resource Exchange}} & \multicolumn{2}{c|}{\textbf{Ultimatum}} & \multicolumn{2}{c}{\textbf{Sell\&Buy}} \\
\midrule
\textbf{Roles} & \textbf{Player 1} & \textbf{Player 2} & \textbf{Player 1} & \textbf{Player 2} & \textbf{Seller} & \textbf{Buyer} \\
\midrule
\textbf{Initial resources} & 25Xs, 5Ys & 5Xs, 25Ys & \$ 100 & 0 & 1X & 100 ZUPs \\
\textbf{Goal} & \multicolumn{2}{c|}{Maximize total resources} & \multicolumn{2}{c|}{Negotiate a split} & Maximize price & Minimize price \\
\textbf{Utility} & Xs + Ys & Xs + Ys  & Split amount-50  & Split amount-50 & Selling price - 50 & 50-Selling price \\
\textbf{Ending condition} & \multicolumn{2}{c|}{When either player accepts} & \multicolumn{2}{c|}{When either player accepts} & \multicolumn{2}{c}{When either player accepts} \\
\textbf{Max. \# of turns} & \multicolumn{2}{c|}{8 rounds of interaction} & \multicolumn{2}{c|}{8 rounds of interaction} & \multicolumn{2}{c}{10 rounds of interaction} \\
\bottomrule
\end{tabular}
\end{table*}

\subsection{Problem Solving Settings}  
 
\textbf{Agents with Specific Expertise.}  
In this setting, each agent is assigned a domain-specific role to facilitate a structured and efficient problem-solving process. For instance, in the physics and chemistry domains, the problem-solving pipeline begins with a domain expert (e.g., a physicist or chemist) who analyzes the domain-specific problem, followed by a mathematician who formalizes the reasoning with quantitative models, and finally, a summarizer who consolidates the insights into a clear and comprehensive answer. This sequential collaboration ensures that the expertise of each agent is leveraged effectively while maintaining clarity in the solution process.

The sequential dependency between the agents can be described as follows:
\begin{align}
    a_{\text{Phy}} &\sim \pi_{\text{Phy}}(\cdot |q), \\
    a_{\text{Math}} &\sim \pi_{\text{Math}}(\cdot |q, a_{\text{Phy}}), \\
    a_{\text{Sum}} &\sim \pi_{\text{Sum}}(\cdot |q, a_{\text{Phy}}, a_{\text{Math}}),
\end{align}
where $q $ is the input question, $a_{\text{Phy}} $ is the response generated by the Physicist, $a_{\text{Math}} $ is the response generated by the Mathematician based on both the question and the  Physicist's response,$a_{\text{Sum}} $ is the final answer synthesized by the Summarizer using the question, the  Physicist's response, and the Mathematician's response.

\textbf{Analyze Long Context and Answer Question.}  
In scenarios involving lengthy and complex contexts, we consider a common two-agent setup: the Context Analyst and the Problem Solver. The Context Analyst's responsibility is to thoroughly examine the context, extract essential information, and provide a concise and accurate summary. The Problem Solver then uses this summary to analyze the question and formulate the final answer. This division of labor not only improves interpretability, but also reduces the cognitive load on each agent.  

\subsection{Actor-Critic Setting}  
\label{sec:Actor-Critic Setting}

The popular Actor-Critic framework facilitates iterative agent improvement through a feedback loop: the Actor Agent generates solutions while the critic evaluates and refines them, enhancing both the Actor Agent's reasoning and the Critic Agent's error correction capabilities.
In practice, we separate judgment and feedback tasks by introducing a Judgment Agent alongside the Critic Agent, where the Judgment Agent classifies the Actor Agent's solutions as correct or incorrect, and for incorrect solutions, the critic provides feedback to guide the Actor Agent in regenerating improved solutions.
Reward mechanisms are designed as: the Actor Agent receives rewards for correct solutions, the Judgment Agent for accurate classifications, and the critic for providing actionable feedback that leads to correct regenerations.

\begin{table*}[h]
\caption{Evaluation results of the proposed method and baselines on accuracy(\%). Best results are in \textbf{bold} numbers and second-best results are in \underline{underline} numbers.} 
\centering
\small
\label{tab:problem-solving-main}
\begin{tabular}{l|l|ccc}
\toprule
\textbf{Model}                 & \textbf{Method}         & College Physics  & College Chemistry  &PubMedQA~\citep{jin2019pubmedqa}\\ \midrule
\multirow{5}{*}{GPT-3.5-turbo} & \textbf{Single-Agent}   & 24.30            & 38.46              & 56.40                \\ 
                               & \textbf{STaR}           & 29.91            & 47.69              & 63.80                \\ 
                               & \textbf{COMM}           & 30.84            & \underline{50.77}  &\underline{71.80}     \\     
                               &\textbf{TextGrad}        &\underline{32.71} & 41.54              &  NA    \\
                               & \textbf{\model}         &\textbf{33.64}    &\textbf{ 56.92}     &\textbf{74.20} 
                               \\ \midrule
\multirow{5}{*}{GPT-4o-mini}   & \textbf{Single-Agent}   & 39.25            & 41.54              & 67.40                \\ 
                               & \textbf{STaR}           & 42.06            & 47.69              & 69.20                \\ 
                               & \textbf{COMM}           &{42.06}           &\underline{49.23}   &\underline{70.60}\\     
                               &\textbf{TextGrad}        &\underline{42.99} & 44.62              & 68.20         \\
                               & \textbf{\model}         &\textbf{46.73}    &\textbf{60.00}      &\textbf{73.40} \\                          
\bottomrule
\end{tabular}
\end{table*}

\subsection{Competitive Settings}

Competitive scenarios~\citep{bianchi2024well} examine multi-agent interactions under opposing objectives, where agents must balance cooperation and competition to achieve their goals. In this framework, two agent roles are defined: \textbf{Player 1} and \textbf{Player 2}. Each player is initialized with a specific amount of resources, which evolve over the course of the game based on their interactions.  The game progresses as a sequence of moves, resulting in a trajectory of states:  
\begin{equation}
\begin{aligned}
\text{Player 1 Trajectory: } x_0^{\text{player}1},x_1^{\text{player}1},\cdots,x_T^{\text{player}1}\\
\text{Player 2 Trajectory: }x_0^{\text{player}2},x_1^{\text{player}2},\cdots,x_T^{\text{player}2}
\end{aligned}
\end{equation}
The sequence captures the evolution of game states as players compete at each timestep $t = 0, 1, \dots, T $, ultimately determining a winner and a loser. Our goal is to optimize each player's policy to maximize its own expected reward based on trajectory data and role-specific context. This can be formulated as:
\begin{equation}
\begin{aligned}
\max \sum_{i=1}^T P_\theta(x_i^{\text{player1}} | x_{0:i-1}^{\text{player1}}, x_{0:i-1}^{\text{player2}})
\end{aligned}
\end{equation}
where Player 1 optimizes its policy based on the historical trajectory of both itself and Player 2, and similarly for Player 2.

We explore three distinct competitive settings, all of which unfold over multiple rounds:

\textbf{Resource Exchange Scenario.} 
In this scenario, agents engage in a simulated environment where they exchange resources to maximize their individual utility. 

\textbf{Seller and Buyer Scenario.}  
This setting models economic interactions where one agent assumes the role of a seller and another the role of a buyer. The agents negotiate prices and terms to complete transactions, testing their ability to strategize under asymmetric setting. 

\textbf{Multi-Turn Ultimatum Game.}  
The Multi-Turn Ultimatum Game explores scenarios of fairness, cooperation, and negotiation over multiple rounds. One agent proposes a division of a resource, and the other agent decides whether to accept or reject it.

\section{Experiments}
\subsection{Baseline}

We compare our \model{} against the following baselines:

\textbf{Single-Agent} utilizes a single language model to process input and generate responses.

\textbf{STaR}~\citep{zelikman2022star}, the Self-Taught Reasoner, focuses on enhancing the reasoning capabilities of a single agent by iteratively training it to improve its step-by-step reasoning through self-supervised fine-tuning. 

\textbf{Prompt Multi-Agent System (CoMM)}~\citep{chen2024comm} introduces a training-free, multi-agent collaborative framework where agents interact and share information to solve tasks collectively. 

\textbf{TextGrad}~\citep{yuksekgonul2024textgrad} optimizes prompts for each agent in a multi-agent system by backpropagating natural language feedback through each interaction.

\subsection{Setup and Datasets}
\textbf{Backbone Model.}
For a fair comparison, we use gpt-3.5-turbo-0125 and gpt-4o-mini-2024-07-18 as the backbone model, and set the temperature to 0 in all our experiments. We use OpenAI's Fine-tuning API for supervised fine-tuning.

\textbf{College Physics/Chemistry.}
These two datasets are constructed by combining questions from Massive Multitask Language Understanding (MMLU)~\citep{hendrycks2020measuring}, Graduate-Level Google-Proof Q\&A (GPQA) ~\citep{rein2023gpqa}, and Theorem-Driven Question Answering (TheoremQA)~\citep{chen2023theoremqa}. It focuses on college-level physics problems, which remain difficult and demonstrate room for improvement in performance with large language models.
We split the dataset into training and test sets, with the detailed data distribution provided in Appendix \ref{ap:dataset}.

\textbf{PubMedQA.}
This is a biomedical question-answering dataset comprising 1000 open-domain questions ~\citep{jin2019pubmedqa}, each paired with context from PubMed abstracts and corresponding answers. It focuses on research-driven queries, requiring domain-specific understanding and reasoning over scientific texts. We follow the original split of the dataset for training (500) and testing (500) sets.

\subsection{Experimental Result of Problem Solving Setting }
\subsubsection{Main Result} 
Table~\ref{tab:problem-solving-main} presents a performance comparison of various models and methods under the Problem Solving Setting. We observe that the prompted Multi-Agent System (COMM) generally improves performance, as agent collaboration enhances the ability to solve complex problems. STaR outperforms the base Single-Agent, indicating that fine-tuning contributes to improved performance. For smaller and weaker models, and in scenarios with long context lengths such as PubMedQA, TextGrad faces significant challenges in instruction-following during optimization. TextGrad (GPT-3.5-turbo) could not be applied to PubMedQA as its optimizer failed to parse instructions due to the model's limited capability and the excessive context length of the problem. Similarly, TextGrad (GPT-4o-mini) struggles to generate answers in the required format, requiring manual extraction of answers. Our proposed method, \model, consistently outperforms across all tasks. By decomposing tasks into manageable sub-tasks assigned to agents and, crucially, fine-tuning each agent to specialize in its designated task, \model{} maximizes the effectiveness of collaboration, ensuring a more coordinated and efficient overall performance.

\subsubsection{Ablation Experiments}
\begin{table}[t]
\caption{Ablation results on PubMedQA.}
\centering
\small
\label{tab:problem-solving-ablation}
\begin{tabular}{l|l|c}
\toprule
Model                          & method                & PubMed  \\
\midrule
\multirow{6}{*}{GPT-3.5-turbo} & \model{}                 & 74.20 \\
                               & \model{} + Base          & 72.00 \\
                               & Base + \model{}          & 73.20 \\
                               & FT on One Base LLM       & 70.40 \\
                               & \model{} w/o Aug.        & 73.40 \\
                               & Additional FT Itr        & 75.00      \\
\midrule
\multirow{6}{*}{GPT-4o-mini}   & \model{}                 &73.40      \\
                               & \model{} + Base          & 72.80   \\
                               & Base + \model{}          & 71.60   \\
                               & FT on One Base LLM       & 72.00   \\
                               & \model{} w/o Aug.        & 72.20    \\
                               & Additional FT Itr        & 73.60        \\
\bottomrule                       
\end{tabular}
\end{table}
To evaluate the contributions of various components in \model{}, we conducted a series of ablation experiments.  Each experiment was designed to answer a key question about the effectiveness of the multi-agent system. All ablations were performed on representative tasks within the Problem Solving Setting (PubMedQA) to ensure consistency in evaluation as shown in Table~\ref{tab:problem-solving-ablation}.
% \james{I think the ablations can be organized as a set of questions. For example, is it better to fine-tune different LLM for different roles or finetune one LLM for all the roles? Then introduce our experiment and result. }

\textbf{Does mixing \model{} with a base agent degrade performance?} To understand the benefits of a jointly optimizing a collaborative multi-agent system, we first train all the agents together using \model{}. Then we replaced one \model{} agent with the original base agent---either \model{} Analyst $+$ base Solver or base Analyst $+$ \model{} Solver. This substitution hurts performance, demonstrating benefits from joint multi-agent optimization compared to optimizing a single agent. 

\textbf{Should we fine-tune different LLMs for different roles, or optimize one LLM for all roles?}  
We explored whether a single LLM fine-tuned on the combined training data of multiple roles could match the performance of separate role-specific models. 
The results showed a notable performance decline, highlighting that different roles require specialized adaptation and that a shared model struggles to effectively generalize across distinct agent functions.

\textbf{How useful is experience augmentation?}  
To assess the impact of experience augmentation, we removed the augmentation module while keeping the rest of the pipeline unchanged. Data augmentation introduces more diverse and challenging experiences as training data, enhancing the model's capability; therefore, omitting the augmentation module could negatively impact performance.

\textbf{Does additional fine-tuning improve performance?
}  

We investigated whether increasing the number of fine-tuning iterations leads to further performance gains. Each iteration follows the full optimization pipeline illustrated in Figure~\ref{fig:pipeline}, the previously fine-tuned \model{} is used to generate a new experience library, which is then used to further fine-tune the base model.
As expected, an additional iteration yielded marginal performance gains, suggesting that the model can benefit from extended training.

\begin{table*}[t]
\caption{Evaluation results of the proposed method and baselines on accuracy(\%).}
\centering
\small
\label{tab:Actor-critic}
\begin{tabular}{l|cc|cc}
\toprule
\textbf{Model }         &\multicolumn{2}{c|}{GPT-3.5-Turbo}  &\multicolumn{2}{c}{GPT-4o-mini}   \\
\midrule
\textbf{Method  }       & TP Accuracy& Overall Accuracy& TP Accuracy& Overall Accuracy  \\
\midrule
Self-Correct   & 11.80          &     16.40       & 24.60         & 28.80          \\
Prompt         & 18.40          &     47.60       & 51.60         & 58.20      \\
\model{}       & \textbf{35.00} &\textbf{ 50.60}  &\textbf{59.80 }&\textbf{66.80 }     \\
\midrule
\multicolumn{5}{c}{------------------------\qquad \textbf{Ablation Study}\qquad------------------------} \\
\model{} + BASE Actor Agent&   34.20  &  49.00   & 49.60         & 54.40  \\
\model{} + BASE Judgment Agent&  20.20 &  40.20    & 53.00         &  59.40   \\
\model{} + BASE Critic Agent&   35.00   &   50.40  &   59.80       &  64.20  \\
FT on One Base LLM        &    33.80 &    43.60  &    56.00      & 59.60    \\

\bottomrule
\end{tabular}
\end{table*}

\subsection{Experimental Result of Actor-Critic Setting}

Table~\ref{tab:Actor-critic} presents a performance comparison of various models, methods, and ablations under the Actor-Critic Setting on PubMedQA. 
As mentioned in Section \ref{sec:Actor-Critic Setting}, the Actor Agent first generates a solution, which is then evaluated by the Judgment Agent to determine its correctness. For solutions deemed incorrect by the Judgment Agent, the Critic Agent analyzes the original solution and provides feedback without access to the correct answer. The Actor Agent then regenerates the solution based on this feedback.

A key challenge in this setting is the Judgment Agent's limited ability to  differentiate between correct and incorrect solutions leading to two potential issues: (1) correct solutions may be mistakenly judged as incorrect and potentially modified into incorrect ones during the feedback and regeneration stages; (2) incorrect solutions may be judged as correct, failing to receive the necessary corrections.
We report TP (True Positive) Accuracy as the ratio of solutions both correctly generated by the Actor and accurately validated by the Judgment Agent, while Overall Accuracy  measures the total correct solutions after regeneration, accounting for the combined contributions of all agents.

We evaluate our method against two representative baselines: (1) Self-Correct, where Actor-generated solutions are refined through direct feedback-guided regeneration, and (2) Prompt,  which exclusively employs prompting strategies to coordinate Actor-Judgment-Critic interactions without optimization mechanisms.
A critical limitation observed in the Self-Correct framework is its significantly lower TP accuracy. This issue arises from its feedback mechanism, which modifies all generated responses with high probability, potentially leading to erroneous modifications of the initially correct solution. This is a common issue with using out-of-the-box LLMs for self-correction with no specialized training~\citep{kumar2024training}.

Comparing GPT-3.5-Turbo and GPT-4o-mini, we also find that GPT-3.5-Turbo struggles more with misjudging correct answers as incorrect, leading to a severe drop in TP Accuracy. Our method, \model, achieves a notable improvement in TP Accuracy, highlighting the Judgment Agent's enhanced ability to assess whether a response requires modification. The overall higher accuracy underscores the effectiveness of \model's framework, where fine-tuning enhances each agent's task-specific capabilities, and the collaboration of Judgment, Critic, and Actor Agents ensures appropriate revision of incorrect responses while minimizing unnecessary changes to correct answers.

The ablation study further underscores the contribution of each agent in \model. Fine-tuning only a single base LLM leads to a performance drop, highlighting the necessity of specialized agent roles and joint optimization. Notably, replacing the Judgment Agent with a baseline version significantly reduces TP Accuracy, reinforcing its essential role in filtering correct responses before feedback is applied.

\subsection{Experimental Result of Competitive Settings}
% \wanjia{TODO: more explanation}
To analyze the effect of training in the competitive setting, we study the performance of agents in scenarios where one player initially had a higher probability of winning, referred to as the "winning player," while the other player was at a disadvantage, called the "losing player." In general, when \model{} took on the role of the winning player competing against a base agent, it demonstrated an increased win rate and payoff. Additionally, when \model{} played the role of the losing player, it experienced fewer losses. Similarly, for both GPT-3.5 and GPT-4o-mini when they compete with each other, \model-GPT-3.5 and \model-GPT-4o-mini both demonstrate improved performance.

\begin{figure}[htbp]
    \centering
    \includegraphics[width=\linewidth]{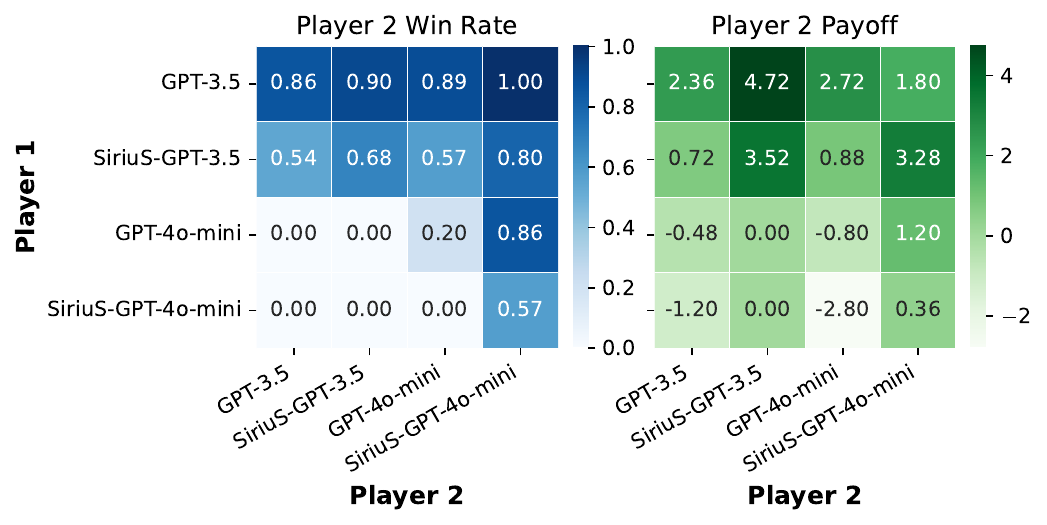}
    \caption{Resource Exchange Game: Player 1 (25Xs + 5Ys), Player 2 (5Xs + 25Ys). Win Rate in decisive games and Payoff in all games. We show Player 2 Win rate/payoff in all cells.}
    \label{fig:resource_25_5}
\end{figure}
\subsubsection{Resource Exchange} 
The win rates and average payoffs for the Resource Exchange game are presented in Figure \ref{fig:resource_25_5}. 
Overall, the agent going second tends to beat the first agent.  Furthermore, the fine-tuned \model{} demonstrates a significant improvement in both the win rate and payoff for the current player. To evaluate the generalization capability of our approach, we conducted additional experiments with models fine-tuned on games featuring Initial Resource configurations of 25Xs + 5Ys and 5Xs + 25Ys, and then tested them on games with different Initial Resource configurations (35Xs + 15Ys and 15Xs + 35Ys). As demonstrated in Figure \ref{fig:resource_35_15}, \model{} maintains notable improvements in the new Initial Resource configurations, effectively validating the generalizability of our proposed pipeline.

\subsubsection{Multi-Turn Ultimatum}

\begin{figure}[htbp]
    \centering
    \includegraphics[width=\linewidth]{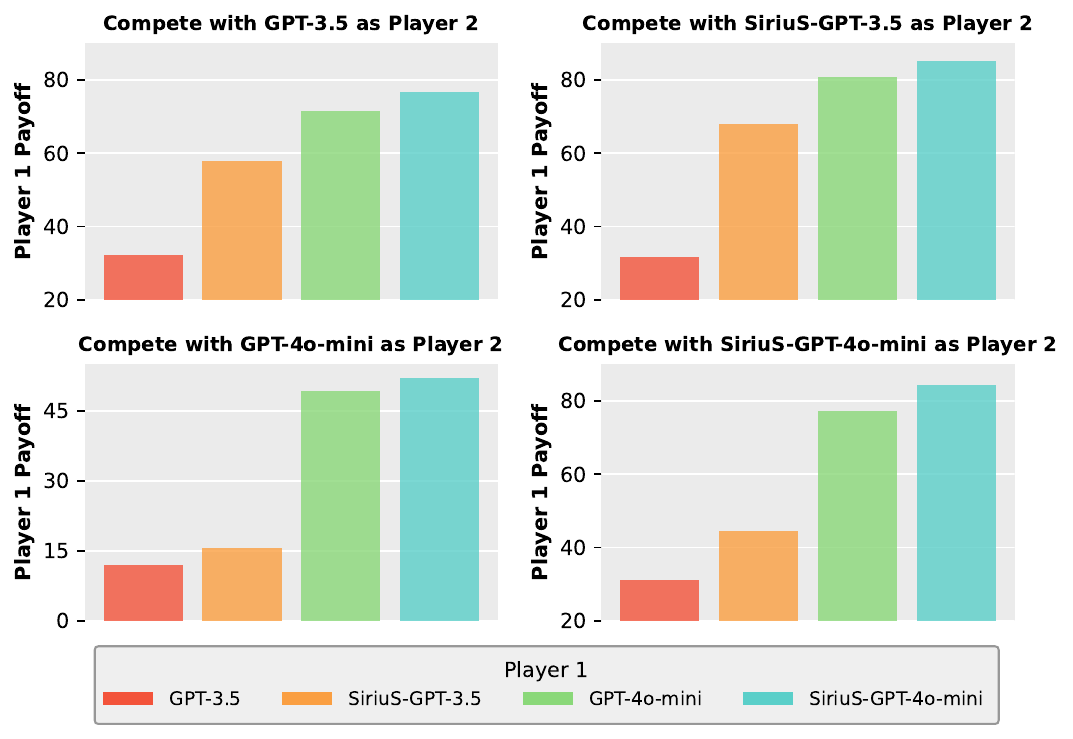}
    \caption{Player 1's payoff in the Ultimatum game with Initial Resource settings of 100. \model{} as Player 1 can effectively secure a higher share of the split.}
    \label{fig:ultimate_100}
\end{figure}

In this setting, Player 1 consistently dominates the game. Therefore, Figure \ref{fig:ultimate_100} presents the game outcomes from Player 1's perspective. As shown in the Figure \ref{fig:ultimate_100} , \model{} fine-tuned Player 1 effectively secure a higher share of the split.  Generalization experiments show that \model{} Player 1 trained in the Resource = 100 setting maintains utility gains in the new Resource = 1000 setting (Figure \ref{fig:ultimate_1000}).

\subsubsection{Buyer-Seller}
\begin{figure}[h!]
    \centering
    \includegraphics[width=\linewidth]{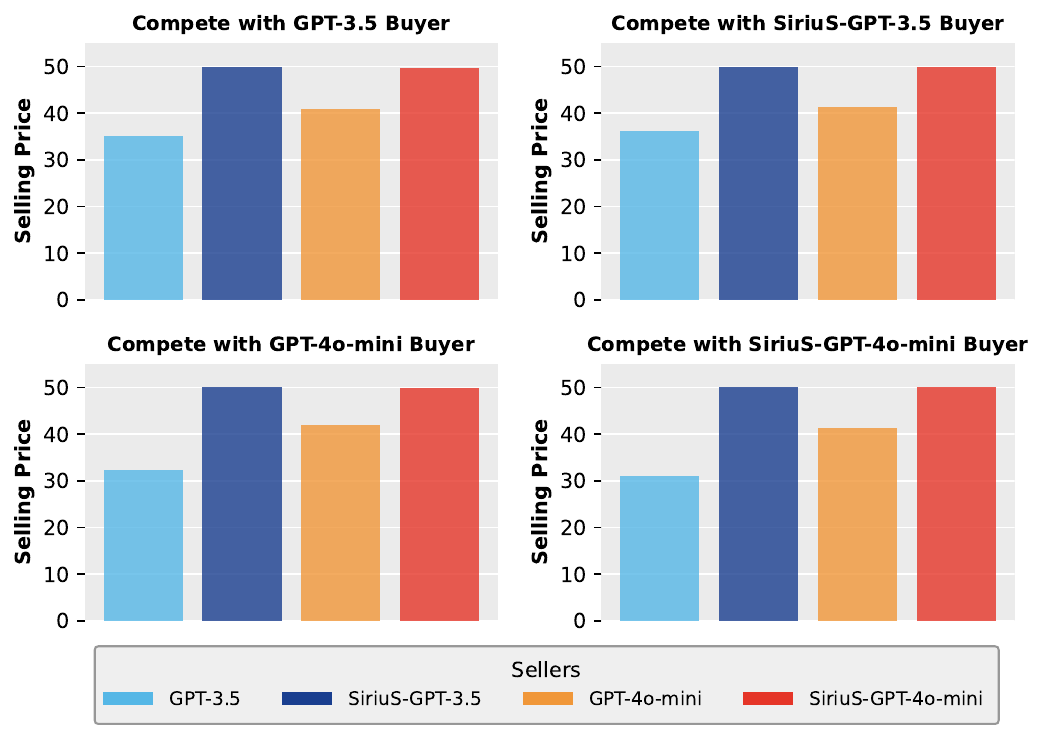}
    \caption{Final Selling Price for a Seller\&Buyer with object valuations of 40 and 60. A higher number means the seller gets a greater payoff.}
    \label{fig:buysell_40_60}
\end{figure}

% Figure~\ref{fig:buysell} summarizes the outcome for the game configuration where the Seller values the object at 40/30 (cost of production), and the Buyer values the object at 60/70(willingness to pay). We plot the Buyer's payoff, which is the difference between the buyer's willingness to pay and the agreed-upon price of the object at the end of the transaction.  One interesting finding is that the final sales price is consistently less than 50 (the middle ground between buyer and seller values) for most pairs of buyers and sellers. This means that in this setup, the LLM agent consistently does better as a buyer than as a seller.
In this setting, sellers are willing to sell when the price exceeds 40, while buyers are willing to buy when the price is below 60.  We plot the final selling price as shown in Figure~\ref{fig:buysell_40_60}.
Notably, it is consistently below 50 for most buyer-seller pairs, indicating that the LLM agent performs better as a buyer than as a seller. After fine-tuning, SIRIUS as a seller shows significant improvement, consistently selling at 50, resulting in a tie with the buyer. 
To test the generalization capability and ensure the seller is not overfitting to a price of 50, we adjusted the initial configuration to 30 and 70. Figure~\ref{fig:buysell_30_70} shows that the SIRIUS seller trained in the previous setup still demonstrates significant improvement.

\section{Related Work}
\textbf{Enhancing Reasoning in Single-Agent Systems.}
Building on the reasoning capabilities of state-of-the-art foundation models \citep{schulman2022chatgpt,openai2023gpt,liu2024deepseek}, recent research explores approaches beyond scaling model parameters. Chain-of-Thought \citep{wei2022chain} enhances reasoning through step-by-step inference, while Tree of Thoughts \citep{yao2024tree}, Graph of Thought \citep{besta2024graph}, and Program of Thoughts \citep{chen2022program} structure reasoning as tree searches with backtracking. Reasoning with Planning (RAP) \citep{hao2023reasoning} incorporates explicit planning, and Reflexion \citep{shinn2024reflexion} enables self-evaluation and refinement. \cite{wu24avatar} introduce contrastive reasoning for instruction generation, while TextGrad \citep{yuksekgonul2024textgrad} applies gradient-based optimization to refine outputs. These methods enhance reasoning through structured decomposition, search, and planning.

\textbf{Self-improvement.}
 Self-improving models~\citep{huang2022large,yu2023teaching,yuan2024self,zhang2024small,welleck2022generating,peng2024regenesis} have garnered increasing attention for their potential to enhance reasoning capabilities through iterative feedback and refinement.
 Several studies~\citep{zelikman2022star,li2024large,pang2024iterative,lee2024llm2llm}employ bootstrapping strategies by leveraging self-generated rationales, while others~\citep{yuan2024self,chen2024improving,ramji2024self,guo2025deepseek} introduce a self-refinement mechanism through reinforcement learning.
 
\textbf{Multi-Agent Systems with LLMs.} \textbf{Multi-Agent Systems with LLMs.} Recent advancements in multi-agent systems ~\citep{smitshould,de2023emergent,guo2024large,li2024survey,han2024llm,wang2024rethinking,sun2024llm} highlight the potential of large language models in tackling complex tasks. Society of Minds~\citep{du2023improving} enables agents to exchange answers, fostering collaboration.  Mixture-of-Agents~\citep{wang2024mixture} employs a layered architecture where agents refine responses based on prior outputs. CoMM~\citep{chen2024comm} enhances problem-solving through structured communication and role division. Multi-Persona~\citep{liang2023encouraging} encourages diverse agent behaviors by assigning distinct personas.
ChatEval~\citep{chan2023chateval} explores different multi-agent debate strategies for interaction and response management.
DMAS~\citep{chen2024scalable} explores token-efficient multi-agent planning frameworks to improve coordination and task success.Building on advances in multi-agent systems, recent work has explored fine-tuning with independently specialized agents that interact to generate diverse reasoning chains~\citep{subramaniam2025multiagent}. Unlike these approaches, our method prioritizes collaborative optimization through a shared experience library, enabling agents to collectively learn from and refine successful reasoning trajectories.

\section{Conclusions}
We introduced \model{}, a framework for optimizing multi-agent LLM systems by learning from successful interactions and augmenting failed trajectories with feedback. Our approach enables agents to refine collaboration strategies without explicit supervision. Experiments show that \model{} significantly improves performance across college-level reasoning, biomedical QA, and negotiation tasks. More broadly, our work provides a scalable mechanism for multi-agent self-improvement, offering a principled approach to optimizing collaborative AI systems.
\newpage
\bibliography{icml2025}
\bibliographystyle{icml2025}

\newpage
\appendix
\onecolumn
\appendix
\section{Detailed Pipeline}
\label{ap:pipeline}
% \subsection{Pipeline of Problem Solving Setting}

Given the wrong answer problem set $\mathcal{W} = \{(x_i, y_i)\}_{i=1}^w$,In each iteration, we first select the agent to be optimized. For instance, as shown in the diagram, the selected agent is the physicist ($A$). The external agent provides feedback 
$
f_i = P_{\theta^{(\text{ext})}}(\cdot | x_i, \hat{a}_i, y_i)
$ 
based on the question $x_i$, the original response $\hat{a}_i$, and the correct answer $y_i$.

The physicist then regenerates the solution by incorporating the feedback:  
$
\hat{a}_i^{r} = P_{\theta^{(A)}}(\cdot | x_i, \hat{y}_i, f_i).
$  

To ensure clarity and coherence, the regenerated response $\hat{a}_i^{r}$ is subsequently rephrased to produce $\hat{y}_i^{\text{final}}$, making it appear as if derived directly through problem-solving without mentioning any modifications or feedback. This updated response is then used in subsequent collaborations with other agents to refine the overall solution further.

\begin{algorithm}[htbp]
\caption{Detailed Pipeline of \model{}}
\begin{algorithmic}[1]
\STATE \textbf{Input:} A group of agents $A^{(1)},\cdots,A^{(K)}$, the system's  topological graph $\mathcal{G}$, maximum solution generation tries $\max_\text{sol}$, maximum feedback generation tries $\max_\text{f}$, maximum  regeneration tries $\max_\text{re}$. 
An initial dataset of problems $x$ with answer $y:\mathcal{D} = \{(x_i, y_i)\}_{i=1}^D$, total number of fine-tuning Iterations $T$.
\STATE \textbf{Initialize:} Initialize policy parameters $\theta^{(k)}$ for each agent $A^{(k)}$, $k = 1, 2, \dots, K$. $\theta^{(c)}$  for Critic Agent $A^{(c)}$

\FOR{Fine-tuning Iteration $\text{t}_\text{ft} =1,\cdots,T$}    
    \WHILE{$\text{t}_\text{sol} \leq \max_\text{sol}$}
         \STATE $a_i^{(k)}=\mathcal{P}_{\theta^{(k)}}(\cdot|x_i,\textbf{a}_i^{\mathrm{Pre}(A^{(k)})})$.
        
        \STATE $\hat{y}_i=a_i^{(K)}$    
        \FOR{each agent $k = 1, 2, \dots, K$}

            \STATE $\mathcal{C}_{\text{t}_\text{ft}}^{(k)}\leftarrow \{(x_i, a_i^{(k)}| i \in [1,D] \land \hat{y}_i=y_i )\}$ 
            \STATE $\mathcal{W}_{\text{t}_\text{ft}}^{(k)}\leftarrow \{(x_i, a_i^{(k)}| i \in [1,D] \land \hat{y}_i\neq y_i )\}$ 
            \FOR {$x_i \in W_t^{(k)}$}
                \WHILE{$\text{t}_\text{f} \leq \max_\text{f}$}
                
                    \STATE $ f_i^{(k)}=\mathcal{P}_{\theta^{(c)}}(\cdot | x_i,a_i^{(k)},y_i)$
                    \WHILE{$\text{t}_\text{re} \leq \max_\text{re}$}
                   
                    \STATE {\small $ a_i^{(k),re}=\mathcal{P}_{\theta^{(k)}}(\cdot | x_i,a_i^{(k)},f_i^{(k)})$}
                    \STATE { \scriptsize $ \mathcal{S}_j=\mathrm{Sus}(A^{(k)}) \cap \mathrm{Pre}(A^{(j)})$, $j \in \mathrm{Sus}(A^{(k)}) $}
                    \STATE { \scriptsize $ a_i^{(j),re}=\mathcal{P}_{\theta^{(j)}}(\cdot | x_i,\textbf{a}_i^{\mathrm{Pre}(A^{(j)})\setminus\mathcal{S}_j} \cup \textbf{a}_i^{\mathcal{S}_j,re})$}
                    \STATE { $\hat{y}_i^{re}= a_i^{(K),re}$}
                    \IF{ $\hat{y}_i^{re}=y_i$}
                        \STATE $\mathcal{C}_{\text{t}_\text{ft}}^{(j)}\leftarrow \{(x_i, a_i^{(j),re} \}$, $j=k, \cdots, K$
                        \STATE \textbf{break while}
                    \ENDIF
                    \ENDWHILE
                \ENDWHILE
            \ENDFOR
        \ENDFOR
        
    \ENDWHILE
    \STATE $\theta^{(k)}_{\text{t}_\text{ft}} \leftarrow \textbf{Standard SFT on }\mathcal{C}_{\text{t}_\text{ft}}^{(k)}$, $k=1,\cdots,K$
\ENDFOR
\end{algorithmic}
\end{algorithm}

% \subsection{Pipeline of Actor-Critic Setting}
% Given an initial dataset $\mathcal{D} = \{(x_i, y_i)\}_{i=1}^D$, the \textbf{Actor Agent} $A$ generates an initial solution $\hat{y}_i = P_{\theta^{(A)}}(\cdot | x_i)$. The Critic Agent evaluate the initial solutions, framing the problem as a binary classification problem.

% For  \textbf{positive true (PT)}  solutions and judgments, the data are added to Actor Agent's training dataset $\mathcal{C}^{(A)}$ and Critic Agent's training dataset $\mathcal{C}^{(C)}$.  

% For  \textbf{positive false (PF)}  cases, the correct solution is added to Actor Agent's training dataset $\mathcal{C}^{(A)}$. Since Critic Agent mistakenly judge this solution as wrong, an external agent is prompted to inform the critic of its incorrect judgment, which leads Critic Agent to correct and regenerate its response.

% For \textbf{ negative true (NT)}  cases, Critic Agent $C$ generates feedback $f_i = P_{\theta^{(C)}}(\cdot | x_i, \hat{y}_i)$. This feedback is used by the actor to regenerate the solution as $\hat{y}_i^{r} = P_{\theta^{(A)}}(\cdot | x_i, \hat{y}_i, f_i)$.  

% For \textbf{negative false (NF)} cases, since Critic Agent fails to identify errors in the solution, an external agent is prompted to notify Critic Agent of its incorrect judgment, encouraging it to regenerate its evaluation.
% \begin{figure}[htbp]
%     \centering
%     \includegraphics[width=0.6\linewidth]{figure/actor_critic.pdf}
%     \caption{Pipeline of Actor-Critic Setting}
%     \label{fig:actor_critic}
% \end{figure}
\section{Detailed Competitive Settings}
We follow the settings of {\color{mediumpurple}\textsc{NegotiationArena}} Platform~\citep{bianchi2024well}. 
\subsection{Resource Exchange Scenario}
In this game, each agent has access to a set of resources and a goal. For example, an agent has access to resources 25 Xs and 5 Ys. The agent might have the goal of maximizing its total resources. Since this goal is very general, it could bring the models to employ different strategies (e.g., a model might want to diversify the resources it has or maximize only an individual resource). Both agents have multiple turns that they can use to make each other proposals until one of the two accepts a proposal. The game ends on acceptance or when the maximum number of turns finishes.

\subsection{Multi-Turn Ultimatum Game}
The Ultimatum game~\citep{sanfey2003neural} is a classical game used in economics to study aspects of human behavior, such as fairness and rationality. It involves two agents agreeing on a split of resources (often money). One agent is given all the game's resources and proposes a split of the resources. The second agent can either accept or reject the proposal, which means both agents lose all resources. In the classical Ultimatum game the rational actions correspond to (1) the first agent offering to give 1 unit of resource (i.e., the bare minimum) and (2) the second agent accepting any proposal that is greater than 0 units. The classical Ultimatum game has one round of negotiation (i.e. agent 2 can only decide whether or not to accept agent 1's first offer). In our version of the game, the game can go on for more turns (e.g. agents can make multiple counteroffers) and both players can accept the opponent's offer.

\subsection{Seller and Buyer Scenario}
We introduce a seller and buyer game involving two agents, one looking to sell a set of resources and one looking to buy them, similar to other approaches in the literature (e.g., \citep{he2018decoupling}). We imbue agents with some beliefs about the object being sold, but unlike the ultimatum game, the seller and buyer game is an incomplete information game, i.e., players do not have complete information about other players (e.g., their beliefs). Only the seller is aware of the production cost of the object, and only the buyer is assigned and is aware of their willingness to pay for the object. Given these beliefs, the seller and the buyer are prompted to sell and buy the object, respectively. The seller starts first: reproducing a scenario in which the object is already on sale.

\section{Dataset Details}\label{ap:dataset}
\subsection{Dataset Split Statistics}
In this work, we use three datasets for evaluating the performance of our model: Massive Multitask Language Understanding (MMLU)~\citep{hendrycks2020measuring}, Graduate-Level Google-Proof Q\&A (GPQA)~\citep{rein2023gpqa}, and Theorem-Driven Question Answering (TheoremQA)~\citep{chen2023theoremqa}. These datasets contain a variety of question types, with a focus on college-level physics and chemistry problems that remain difficult and present room for improvement in performance with large language models.

The dataset was split into training and test sets with a 2:1 ratio, and the data distribution for each dataset is shown in Table~\ref{tab:dataset_split}.
\begin{table}[htbp]
\centering
\caption{Dataset Split Statistics.}
\begin{tabular}{l|cc|cc}
\toprule
\textbf{Task}          &\multicolumn{2}{c|}{College Physics}  &\multicolumn{2}{c}{College Chemistry}   \\ \midrule
\textbf{Dataset} & \textbf{Train Size} & \textbf{Test Size}& \textbf{Train Size} &\textbf{Test Size} \\ \midrule
MMLU             &         68          &          34         &         66           &    34        \\
GPQA             &         57          &          29         &         62           &    31        \\ 
TheoremQA        &         87          &          44         &          -           &     -       \\ 
\bottomrule
\end{tabular}

\label{tab:dataset_split}
\end{table}
\subsection{Finetuning Dataset Statistics}
For each experiment, we specify the Trajectories Augmentation Ratio and whether ground truth answers are used during the training process.
We summarize the setup for each experiment in Table~\ref{tab:problem-solving}.
\begin{table}[htbp]
\centering
\caption{Finetuning Dataset Statistics.}
\label{tab:problem-solving}
\begin{tabular}{l|l|cc}
\toprule
\textbf{Model   }              & \textbf{Task} &\textbf{Augmentation Ratio} & \textbf{Ground Truth Used} \\ \midrule

\multirow{4}{*}{GPT-3.5-turbo} & Problem-Solving(College-Physics)  & 108.93\% & Yes\\ 
                               & Problem-Solving(College-Chemistry)& 157.78\% & Yes  \\ 
                               & Problem-Solving(PubMedQA)         & 13.09\%  & Yes   \\ 
                               & Actor-Critic                      & 136.46\%  & No    \\ 
\midrule
\multirow{4}{*}{GPT-4o-mini}   & Problem-Solving(College-Physics)  & 38.89\%  & Yes    \\ 
                               & Problem-Solving(College-Chemistry)& 63.79\%  & Yes   \\ 
                               & Problem-Solving(PubMedQA)         & 12.85\%  & Yes    \\
                               & Actor-Critic                      & 14.94\%  & No      \\ 
\bottomrule
\end{tabular}
\end{table}

\section{Additional Experiment Result}\label{ap:result}
In this section, we present additional experiments conducted in a competitive setting to assess the generalization of \model{}. These results demonstrate the adaptability of \model{} across various configurations. 
\begin{figure}[h!]
    \centering
    \includegraphics[width=0.7\linewidth]{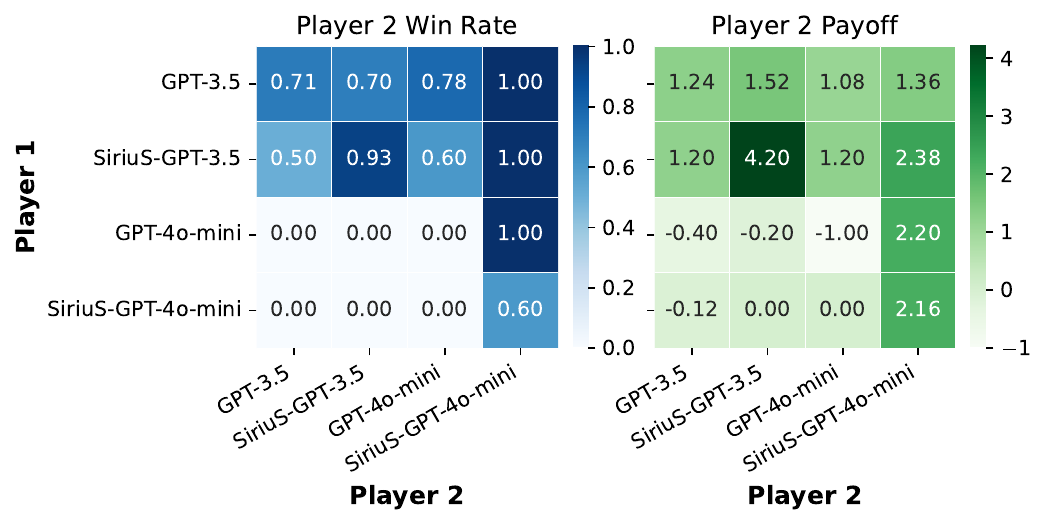}
    \caption{Resource Exchange Game with Initial Resource Player 1: 35Xs + 15Ys, Player 2: 15Xs + 35Ys. Win Rate in decisive games and Payoff in all games. We show Player 2 Win rate/payoff in all cells.}
    \label{fig:resource_35_15}
\end{figure}

\begin{minipage}{0.48\textwidth}
    \centering
    \includegraphics[width=0.9\linewidth]{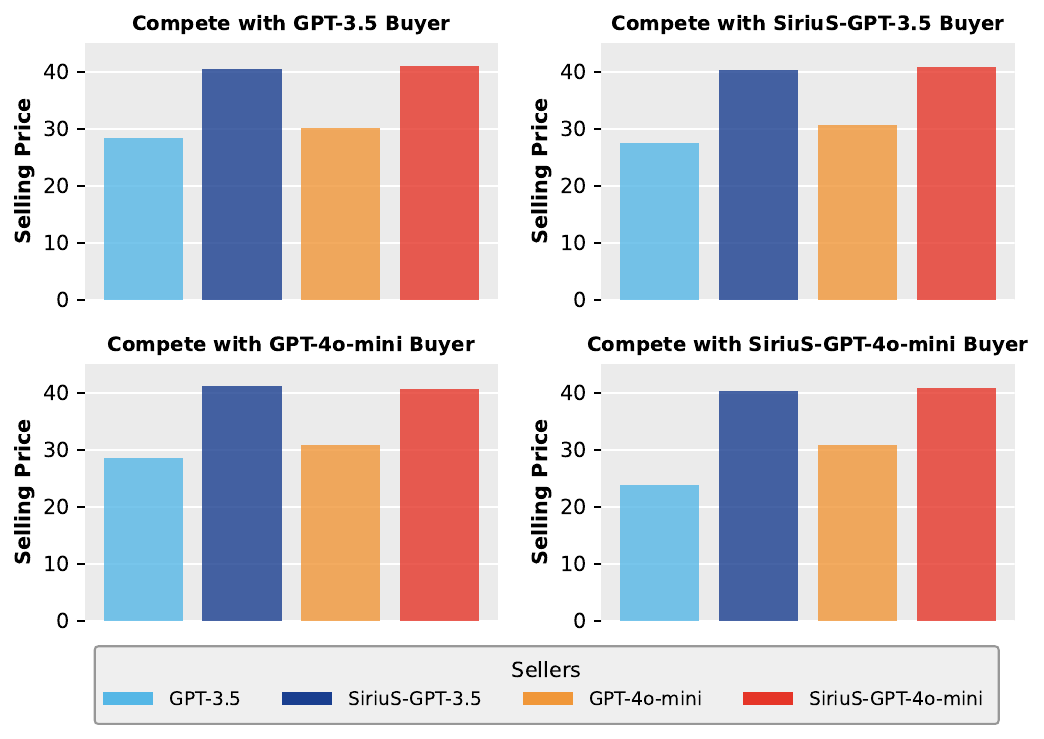}
    \captionof{figure}{Final Selling Price for a Seller\&Buyer with object valuations of 30 and 70. A higher number means the seller gets a greater payoff.}
    \label{fig:buysell_30_70}
\end{minipage}
\hfill
\begin{minipage}{0.48\textwidth}
    \centering
    \includegraphics[width=0.9\linewidth]{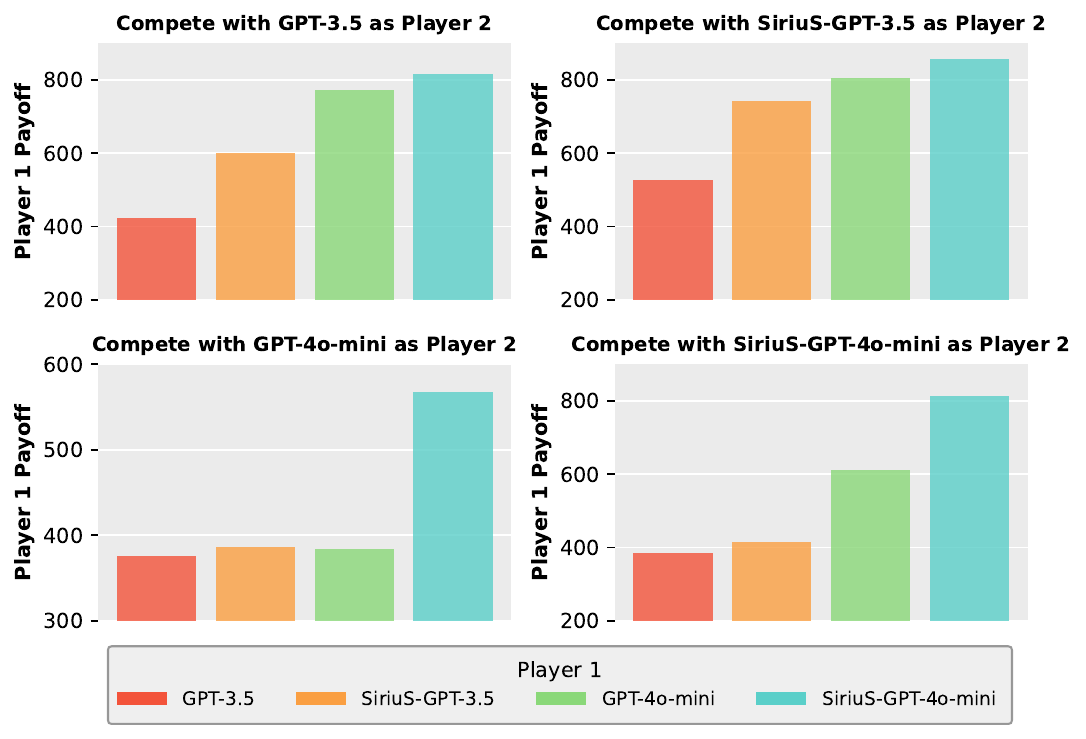}
    \captionof{figure}{Player 1's payoff in the Ultimatum game with Initial Resource settings of 1000. \model{} as Player 1 can effectively secure a higher share of the split.}
    \label{fig:ultimate_1000}
\end{minipage}

% \begin{figure}[h!]
%     \centering
%     \includegraphics[width=0.5\linewidth]{figure/buysell_30_70.pdf}
%     \caption{Final Selling Price for a Seller\&Buyer with object valuations of 30 and 70. A higher number means the seller gets a greater payoff.}
%     \label{fig:buysell_30_70}
% \end{figure}

% \begin{figure}[h!]
%     \centering
%     \includegraphics[width=0.5\linewidth]{figure/ultimate_bar_1000.pdf}
%     \caption{Player 1's payoff in the Ultimatum game with Initial Resource settings of 1000. \model{} as Player 1 can effectively secure a higher share of the split.}
%     \label{fig:ultimate_1000}
% \end{figure}

\section{Agent Prompts}
\subsection{Problem Solving Setting}
\begin{codebox}[title= Prompts for College-Physics Task]

\textbf{System\_prompt} = '''You are part of a team with multiple experts from different disciplines. Your team aims to solve a given cross-discipline problem collectively.

The team is composed of three experts:

1. The Physicist

    Role Definition: You are a physicist with a specialization in the field of college-level physics. Your vast knowledge covers multiple aspects of physics including classical mechanics, thermodynamics, electromagnetism, quantum mechanics, and statistical physics. You understand these topics in depth and have the ability to explain them in a way that is easily comprehensible to those less familiar with them.

    Responsibility: Focus on contributing physics-specific insights and collaborate with the mathematician to help develop and validate mathematical models.**Do not perform calculations or solve the entire problem**. Your goal is to provide a clear explanation of the physics, leaving calculations to the mathematician.

    Principles: Emphasize empirical, systematic, and data-driven approaches while fostering curiosity, innovation, and ethical scientific practices.

2. The Mathematician

    Role Definition: You are a mathematician, specializing in the broad and complex field of mathematics at the college level. Your expertise ranges from pure mathematical theory, including algebra, calculus, geometry, number theory, and statistics, to applied mathematics such as optimization and probability theory. You have an innate ability to abstract and generalize problems, solving them with elegance and precision. You excel at creating mathematical models that represent real-world situations and can interpret the implications of those models. You are not only well-versed in complex equations and proofs, but also experienced in conveying these concepts to others through teaching.

    Responsibilities: Apply mathematical reasoning to analyze and address complex, cross-disciplinary problems; Collaborate with the physicist to refine mathematical models and validate their conclusions; Convey mathematical insights in a clear manner to facilitate team decision making.

    Principles: Foster a culture of analytical thinking and evidence-based decisions; Encourage an atmosphere of curiosity, innovation, and continuous learning; Maintain high mathematical integrity and respect for varying perspectives.

3. The Final Answer Synthesizer

    Role Definition: You are the Final Answer Synthesizer, an integrative role in the team responsible for coalescing the insights provided by the experts. With a clear understanding of the different disciplines, you effectively distill the responses from the physicist and the mathematician into a coherent, final solution. Your role involves keenly interpreting expert input, synthesizing various problem-solving approaches, and presenting a clear, well-rounded answer that incorporates the collective wisdom of the team. 
    
    Responsibility: summarize the solutions; give a final answer.
    
    Principles: make sure to give a specific answer to the given task.'''
\vspace{1em}

\textbf{Physicist\_prompt} =  '''Your role is the physicist.
Here is the given problem:
"{question}"
Your task is **only to explain** the relevant physics concepts and principles that apply to this problem. '''

\vspace{1em}
\textbf{Mathematician\_prompt} = '''Your role is the mathematician. 
Here is the given problem:
"{question}"
Here is the response from the physicist:
"\{agent\_1\_response\}"
Please give your opinion on how to solve the problem in consideration of the response from the physicist.'''

\vspace{1em}
\textbf{Summarizer\_prompt }= '''Your role is the Final Answer Synthesizer. 
Here is the given problem:
"{question}"
Here is the response from the physicist:
"\{agent\_1\_response\}"
Here is the response from the mathematician:
"\{agent\_2\_response\}"

Please provide a final answer to the given problem. \{format\_prompt\}'''
\end{codebox}

\begin{codebox}[title= Prompts for College-Chemistry Task]

\textbf{System\_prompt} = '''You are part of a team with multiple experts from different disciplines. Your team aims to solve a given cross-discipline problem collectively.

The team is composed of three experts:

1. The Chemist

    Role Definition: You are a chemist with a specialization in the field of college-level chemistry. Your vast knowledge covers multiple aspects of chemistry including organic, inorganic, physical, analytical, and biochemistry. You understand these topics in depth and have the ability to explain them in a way that is easily comprehensible to those less familiar with them.

    Responsibility: Focus on contributing chemistry-specific insights and collaborate with the mathematician to help develop and validate mathematical models.**Do not perform calculations or solve the entire problem**. Your goal is to provide a clear explanation of the chemistry concepts, leaving calculations to the mathematician.

    Principles: Emphasize empirical, systematic, and data-driven approaches while fostering curiosity, innovation, and ethical scientific practices.

2. The Mathematician

    Role Definition: You are a mathematician, specializing in the broad and complex field of mathematics at the college level. Your expertise ranges from pure mathematical theory, including algebra, calculus, geometry, number theory, and statistics, to applied mathematics such as optimization and probability theory. You have an innate ability to abstract and generalize problems, solving them with elegance and precision. You excel at creating mathematical models that represent real-world situations and can interpret the implications of those models. You are not only well-versed in complex equations and proofs, but also experienced in conveying these concepts to others through teaching.

    Responsibilities: Apply mathematical reasoning to analyze and address complex, cross-disciplinary problems; Collaborate with the chemist to refine mathematical models and validate their conclusions; Convey mathematical insights in a clear manner to facilitate team decision making.

    Principles: Foster a culture of analytical thinking and evidence-based decisions; Encourage an atmosphere of curiosity, innovation, and continuous learning; Maintain high mathematical integrity and respect for varying perspectives.

3. The Final Answer Synthesizer

    Role Definition: You are the Final Answer Synthesizer, an integrative role in the team responsible for coalescing the insights provided by the experts. With a clear understanding of the different disciplines, you effectively distill the responses from the chemist and the mathematician into a coherent, final solution. Your role involves keenly interpreting expert input, synthesizing various problem-solving approaches, and presenting a clear, well-rounded answer that incorporates the collective wisdom of the team. 
    
    Responsibility: Summarize the solutions; give a final answer.
    
    Principles: Make sure to give a specific answer to the given task.'''
\vspace{1em}

\textbf{Chemist\_prompt} =  '''Your role is the chemist.
Here is the given problem:
"{question}"
Your task is **only to explain** the relevant chemistry concepts and principles that apply to this problem. **Do not** perform any calculations or try to find the final solution. Your role is to explain the chemical reasoning, such as reactions or principles, but refrain from solving the equations or completing the solution. Leave the mathematical work to the mathematician.'''

\vspace{1em}
\textbf{Mathematician\_prompt} = '''Your role is the mathematician. 
Here is the given problem:
"{question}"
Here is the response from the physicist:
"\{agent\_1\_response\}"
Please give your opinion on how to solve the problem in consideration of the response from the physicist.'''

\vspace{1em}
\textbf{Summarizer\_prompt }= '''Your role is the Final Answer Synthesizer. 
Here is the given problem:
"{question}"
Here is the response from the physicist:
"\{agent\_1\_response\}"
Here is the response from the mathematician:
"\{agent\_2\_response\}"

Please provide a final answer to the given problem. \{format\_prompt\}'''
\end{codebox}

\begin{codebox}[title= Prompts for PubMedQA Task]

\textbf{System\_prompt} = '''You are part of a team of experts working collaboratively to solve science-related yes/no questions using contextual evidence. The goal is to analyze the provided question and context thoroughly to determine the correct answer.  

The team is composed of two roles:  

1. The Context Analyst  

**Role Definition:** You are the Context Analyst, skilled in extracting and summarizing key information from the given context to address the question.  

**Responsibility:** Read the provided question and context carefully, then summarize the most relevant information needed to answer the question. Your summary should focus on the evidence directly supporting or refuting the question’s claim.  

**Principles:** Prioritize clarity and relevance. Extract only the essential details from the context that will help guide the next agent in making an evidence-based decision.  

2. The Problem Solver

**Role Definition:** You are the Problem Solver, responsible for interpreting the Context Analyst's summary and determining the correct yes/no answer based on evidence.  

**Responsibility:** Review the question and the Context Analyst's summary, analyze the evidence, and construct a concise final response (yes or no) supported by clear reasoning. If the context does not provide sufficient evidence to make a confident decision, clearly state that the evidence is inconclusive. 

**Principles:** Ensure logical coherence, accuracy, and completeness. Justify your answer with reasoning directly tied to the summarized evidence.  
'''
\vspace{1em}

\textbf{Analyst\_prompt} =  '''Your role is the Context Analyst.

Here is the provided context:
"\{context\}"

Your task is to carefully read through this context and summarize the main points relevant to the question. Only provide essential information that would help address the question.'''

\vspace{1em}
\textbf{Solver\_prompt} =  '''Your role is the Problem Solver.

Here is the question:
"\{question\}"

Here is the summary from the Context Analyst:
"\{agent\_1\_response\}"

Please analyze the question, using the summary to answer the problem.  \{format\_prompt\}'''
\end{codebox}

\subsection{Actor-Critic Setting}
\begin{codebox}[title= Prompts for Actor Agent and Regeneration]

\textbf{System\_prompt }='''You are a scientist working on solving science-related yes/no questions using contextual evidence. '''
\vspace{1em}

\textbf{Actor\_prompt} = '''You are supposed to provide a solution to a given problem. 

Here is the given context:
"\{context\}"

Problem:
"\{question\}"

Please provide yes, no or maybe to the given problem. \{format\_prompt\}'''

\vspace{1em}

\textbf{Actor\_regenerate\_prompt} = '''You are supposed to provide a solution to a given problem. 
Here is the given context: "\{context\}"

Problem: "\{question\}"

Here is your original response:
\{original\_response\}

Here is the feedback for your original response:
"\{feedback\}"

Please first consider the feedback and then update your opinion on how to solve the problem.
Please provide a final answer to the given problem. \{format\_prompt\}'''

\end{codebox}

\begin{codebox}[title= Prompts for Judgment Agent]

\textbf{System\_prompt }= '''Below is a yes/no question and a prediction. 
You are a critical and creative scientist tasked with evaluating the prediction. Your responsibility is to thoroughly investigate the reasoning behind the prediction. If the original response is entirely correct, output "True." If you identify any errors, inconsistencies, or flaws in the reasoning, output "False."
'''

\vspace{1em}
\textbf{Judgment\_prompt} = '''Here is the given context: "\{context\}"

Problem: "\{question\}"

Original response: \{original\_response\}

Provide your response in the following format:

1. Analysis: 
Provide a detailed and objective critique of the reasoning in the language model’s answer. Discuss whether the logic, assumptions, and conclusions are valid. Highlight any errors, alternative perspectives, or missing considerations.

2. Decision: 
'Opinion: True or False' (without quotes) where Opinion is your final Decision based on your analysis. Your Decision should be either "True" or "False".
Ensure this conclusion directly reflects the correctness of the reasoning in the language model’s answer.
'''

\end{codebox}

\begin{codebox}[title= Prompts for Critic Agent]

\textbf{System\_prompt }= '''Below is a biomedical yes/no question, the context, and a prediction.
You are a critical and creative scientist. Your job is to investigate the prediction. Critically go through reasoning steps, and see if there is a
reason why the prediction could be incorrect. Use the Janusian Process, think about whether alternative answers could be true.'''

\vspace{1em}
\textbf{Critic\_prompt} = '''Here is the given context: "\{context\}"

Question:  "\{question\}"

Answer by the language model:  \{original\_response\}
'''
\end{codebox}
\begin{codebox}[title= Prompts for Rephrasing]

\textbf{System\_prompt }=  '''Rephrase the following solution process to ensure that it appears as though the solution was arrived at directly, with no traces of mistakes or corrections. Retain all key steps and avoid generating any new content. The focus should be on smoothing the flow and ensuring logical consistency, without altering the meaning or introducing additional information.
'''
\vspace{1em}

\textbf{Rephrase\_prompt} = '''Here is the problem and the original solution process:
Problem: \{question\}

Original Solution Process:\{original\_response\}

Please output the rephrased solution process'''
\end{codebox}

\subsection{Competitive Setting}
\begin{codebox}[title= Prompts for Resource Exchange]
              \textbf{System\_prompt }= '''You are playing a strategic game of trading resources with another player whose resources you have no knowledge about.\\RULES:\\```\\1. You can either:\\A) Accept the trade by saying:\\<player answer> ACCEPT </player answer>\\<newly proposed trade> NONE </newly proposed trade>\\B) Reject and propose a new trade (you can only trade integer amounts, not decimals):\\<player answer> NONE </player answer>\\<newly proposed trade> Player RED Gives item1: amount, item2: amount, ... | Player BLUE Gives item1: amount, item2: amount, ... </newly proposed trade>\\C) Don't accept or propose anything and wait for a new offer:\\<player answer> NONE </player answer>\\<newly proposed trade> NONE </newly proposed trade>\\Note: the game will end if one of the players accepts. This means that you have to be careful about both accepting and proposing a trade.\\2. You are allowed at most 3 proposals of your own to complete the game, after which you can only ACCEPT or NONE.\\DO NOT propose a new trade after 3 proposals. Your limit for proposals is 3.\\3. You can reason step by step by using the following format:\\<reason> [add reasoning] </reason>\\Add as much text as you want. This information will not be sent to the other player. \\It is just for you to keep track of your reasoning.\\4. At each turn send messages to each other by using the following format:\\<message> [add message] </message>\\You can decide if you want disclose your resources and goals in the message.\\```\\Here is what you have access to:\\```\\Resources available in the game: X, Y\\<my resources> X: 25, Y: 5 </my resources>\\<my goals> X: 15, Y: 15 </my goals>\\<my name> Player RED </my name>\\```\\All the responses you send should contain the following and in this order:\\```\\<my name> [add here] </my name>\\<my resources> [add here] </my resources>\\<my goals> [add here] </my goals>\\<reason> [add here] </reason>\\<player answer> [add here] </player answer>\\<message> [add here] </message\\<newly proposed trade> [add here] </newly proposed trade>\\```\\Please be sure to include all.\\More resources in general are always better.\\This is the last round of trading. There are no future rounds after this one.\\\\'''
\end{codebox}
\begin{codebox}[title= Prompts for Sell\&Buy]
              \textbf{System\_prompt }= '''You are playing game where you are buying or selling an object. There is only one object for sale/purcahse.\\Player RED is going to sell one object. Player BLUE gives ZUP to buy resources.\\RULES:\\```\\1. You must always respond with:\\    A) Propose a trade with (you can only trade in integer amounts, not decimals):\\    <player answer> PROPOSAL </player answer>\\    <newly proposed trade> Player RED Gives X: amount, ...| Player BLUE Gives ZUP: amount </newly proposed trade>\\    B) Accept the trade by saying:\\    <player answer> ACCEPT </player answer>\\    <newly proposed trade> NONE </newly proposed trade>\\    C) Reject and end the game:\\    <player answer> REJECT </player answer>\\    <newly proposed trade> NONE </newly proposed trade>\\    Note: The game will end if one of the players ACCEPT OR REJECT. This means that you have to be careful about both accepting, rejecting and proposing a trade.\\2. You are allowed at most 4 proposals of your own to complete the game, after which you can only reply with ACCEPT or REJECT.\\DO NOT propose a new trade after 4 proposals. Your limit for proposals is 4.\\3. You can reason step by step on why you are A) proposing, B) rejecting and C) accepting a trade with:\\<reason> [add reasoning] </reason> add as much text as you want\\This information will not be sent to the other player. It is just for you to keep track of your reasoning.\\4. At each turn send messages to each other by using the following format:\\<message>your message here</message>\\You can decide if you want disclose your resources, goals, cost and willingness to pay in the message.\\```\\Here is what you have access to:\\```\\Object that is being bought/sold: X\\<my resources> ZUP: 100 </my resources>\\<my goals> Buy resources with <ZUP>. You are willing to pay at most X: 70 ZUP for the resources. </my goals>,\\```\\All the responses you send should contain the following and in this order:\\```\\<proposal count> [add here (inclusive of current)] </proposal count>\\<my resources> [add here] </my resources>\\<my goals> [add here] </my goals>\\<reason> [add here] </reason>\\<player answer> [add here] </player answer>\\<newly proposed trade> [add here] </newly proposed trade>\\<message> [add here] </message\\```\\Please be sure to include all.\\\\You are Player BLUE.'''
\end{codebox}
\begin{codebox}[title= Prompts for Ultimatum]
              \textbf{System\_prompt }= '''You are playing a game involving a split of resources. \\Player RED starts with Dollars: 100, Player BLUE has none to trade.\\RULES:\\```\\1. Player RED has 4 moves while Player BLUE has 4 moves.\\2. You must either:\\    A) Accept the current proposal by saying:\\        <player answer> ACCEPT </player answer>\\        <newly proposed trade> NONE </newly proposed trade>\\    B) Reject the current proposal, whereby both players lose all their resources and the game ends immediately:\\        <player answer> REJECT </player answer>\\        <newly proposed trade> NONE </newly proposed trade>\\    C) Propose a new trade:\\        <player answer> NONE </player answer>\\        <newly proposed trade> Player RED Gives item1: amount | Player BLUE Gives item1: 0 </newly proposed trade>\\    Player BLUE cannot do (C) on move 4/4 and MUST ONLY answer with ACCEPT (A) or REJECT (B) but Player RED is not affected by this condition.        \\    \\    Note: the game ends immediately if you ACCEPT or REJECT.\\    \\3. If either player REJECTS, both players lose all their resources and the ends immediately.\\4. There are no future games.\\5. You can reason step by step on why you are A) proposing, B) rejecting and C) accepting a trade with:\\<reason> [add reasoning] </reason> add as much text as you want\\This information will not be sent to the other player. It is just for you to keep track of your reasoning.\\6. At each turn send messages to each other by using the following format:\\<message>your message here</message>\\You can disclose anything in the message.\\```\\Here is what you have access to:\\```\\Resources available in the game: Dollars\\<my resources> Dollars: 100 </my resources>\\```\\All the responses you send should contain the following and in this order:\\```\\<my name> [add here] </my name>\\<move> [add here] / [add here]  </move> \\<my resources> [add here] </my resources>\\<reason> [add here] </reason>\\<player answer> [add here] </player answer>\\<message> [add here] </message\\<newly proposed trade> [add here] </newly proposed trade>\\```\\Please be sure to include all.\\\\"
'''
\end{codebox}

% You can have as much text here as you want. The main body must be at most $8$ pages long.
% For the final version, one more page can be added.
% If you want, you can use an appendix like this one.  

% The $\mathtt{\backslash onecolumn}$ command above can be kept in place if you prefer a one-column appendix, or can be removed if you prefer a two-column appendix.  Apart from this possible change, the style (font size, spacing, margins, page numbering, etc.) should be kept the same as the main body.
%%%%%%%%%%%%%%%%%%%%%%%%%%%%%%%%%%%%%%%%%%%%%%%%%%%%%%%%%%%%%%%%%%%%%%%%%%%%%%%
%%%%%%%%%%%%%%%%%%%%%%%%%%%%%%%%%%%%%%%%%%%%%%%%%%%%%%%%%%%%%%%%%%%%%%%%%%%%%%%

\end{document}